\documentclass[conference]{IEEEtran}
\IEEEoverridecommandlockouts

\usepackage{cite}
\usepackage{amsmath,amssymb,amsfonts}
\usepackage{algorithmic}
\usepackage{graphicx}
\usepackage{textcomp}
\usepackage{xcolor}
\usepackage{booktabs}
\def\BibTeX{{\rm B\kern-.05em{\sc i\kern-.025em b}\kern-.08em
    T\kern-.1667em\lower.7ex\hbox{E}\kern-.125emX}}

\usepackage{array}
\usepackage{tabularx}
\usepackage{enumitem}
\usepackage{float}
\usepackage[most]{tcolorbox}

\begin{document}

\title{Select or Rewrite: Oracle-Guided Fine-Tuning for Generative ASR Post-Editing from N-Best Hypotheses\\
}

\author{
\IEEEauthorblockN{
Shiran Liu\textsuperscript{1,*,\dag},
Gaohui Xu\textsuperscript{2,*},
Sergio Burdisso\textsuperscript{1},
Thibault Ba\~{n}eras-Roux\textsuperscript{1},
Shashi Kumar\textsuperscript{1},
Petr Motlicek\textsuperscript{1,3}}
\IEEEauthorblockA{
\textsuperscript{1}\textit{Idiap Research Institute}, Martigny, Switzerland\\
\textsuperscript{2}\textit{KTH Royal Institute of Technology}, Stockholm, Sweden\\
\textsuperscript{3}\textit{Brno University of Technology}, Brno, Czech Republic\\
shiran.liu@foxmail.com, gaohui@kth.se\\
\textsuperscript{*}Equal contribution. \quad \textsuperscript{\dag}Corresponding author.}
}

\newenvironment{promptbox}{
  \begin{tcolorbox}[
    colback=black!1,
    colframe=black!25,
    boxrule=0.4pt,
    arc=2mm,
    left=2mm,right=2mm,top=1mm,bottom=1mm,
  ]
  \ttfamily\footnotesize
  \setlength{\tabcolsep}{0pt}
  \renewcommand{\arraystretch}{1.15}
}{
  \end{tcolorbox}
}

\maketitle

\begin{abstract}
Large language models can post-edit ASR by using linguistic context to compare N-best hypotheses and rewrite the transcript when needed, often improving WER without retraining the recognizer. However, prompt-only correction is brittle: it may over-edit when top-1 is correct and hallucinate when the reference is absent, and often needs very large models. We propose oracle-guided fine-tuning that turns N-best lists into supervision via three curated pools: corrections (reference in the list but not top-1) to teach selection, regularizers (top-1 correct) to discourage unnecessary rewrites, and hard negatives (reference absent) to suppress hallucination.
Fine-tuning on this data provides more effective supervision than random subset fine-tuning for learning a hybrid select-or-rewrite policy, with the clearest gains from around 3B parameters on the proprietary dataset but less conclusive results on CommonVoice, a divergence suggesting that public read-speech text may overlap with LLM pretraining and mask genuine post-editing gains.
\end{abstract}

\begin{IEEEkeywords}
automatic speech recognition, ASR post-editing, N-best rescoring, large language models, oracle-guided fine-tuning. 
\end{IEEEkeywords}

\section{Introduction}
End-to-end automatic speech recognition (ASR) systems such as Whisper \cite{radfordRobustSpeechRecognition2023} have substantially improved transcription quality, yet errors remain frequent on challenging audio and long-tail phenomena (e.g., proper nouns, numbers, and rare word forms). Crucially, many of these errors may be recoverable: beam-search decoding produces an N-best list that often contains a more accurate alternative than the top-1 output. This naturally motivates a second-pass module that uses broader linguistic context to compare candidates and resolve ambiguities that local, token-level scoring may miss.

Classic second-pass rescoring typically combines acoustic and language model scores through linear interpolation (e.g., shallow fusion) \cite{gulcehreUsingMonolingualCorpora2015}. While efficient, such methods still tend to evaluate hypotheses largely independently and can underutilize the confusion structure encoded in the N-best list, limiting gains from long-range semantic disambiguation \cite{shivakumarSpeechRecognitionRescoring2024}. Recent work therefore explores large language models (LLMs) as post-editors or rerankers. In principle, prompted LLMs can reason about candidate differences, but practical use remains brittle: unconstrained generation may hallucinate or over-edit when the Top-1 hypothesis is already correct, while restricting outputs to the N-best list caps performance at the oracle candidate quality \cite{maNbestT5Robust2023, maCanGenerativeLarge2023}. Moreover, strong results often rely on very large or multimodal models \cite{shivakumarSpeechRecognitionRescoring2024}, increasing latency and cost; smaller LLMs frequently underperform without task-specific adaptation.

We address these issues with Oracle-Guided Fine-Tuning, a data-centric framework that teaches an LLM a hybrid \textbf{select-or-rewrite} policy for ASR post-editing from N-best hypotheses. The key idea is to use oracle information during training (i.e., the ground-truth transcript) to label how each N-best list should be handled, and to curate a balanced instruction-tuning set that explicitly trains three behaviors: (i) selection when the correct transcript is present in the list, (ii) conservatism when Top-1 is already correct (to discourage unnecessary edits), and (iii) hallucination suppression when the reference is absent (to prevent unsupported rewrites). This turns N-best lists into supervision that aligns the model with realistic second-pass decision boundaries.


Across two datasets with contrasting properties, we evaluate whether oracle-guided curation provides better supervision than randomly selected training examples under the same fine-tuning data budget. Rather than a uniform gain, we find a clear divergence between the two settings. On the proprietary conversational corpus, the curated subset is clearly the most effective and its benefit becomes reliable from around 3B LLM parameters; on CommonVoice, however, its advantage over random selection and prompting is less conclusive. We argue that this contrast is itself informative: it points to the influence of domain and, possibly, of overlap between public read-speech benchmark text and LLM pretraining data, motivating our discussion of how such factors complicate the evaluation of LLM-based ASR post-editing on public benchmarks.

\textbf{Contributions.}
(1) We propose an oracle-guided fine-tuning recipe that converts ASR N-best lists into curated supervision for a select-or-rewrite post-editor.
(2) We introduce a three-pool data construction strategy, corrections, regularizers, and hard negatives, that targets over-editing and hallucination failure modes.
(3) We show that oracle-guided curation outperforms random subset fine-tuning under a matched data budget, especially on the proprietary definedAI dataset and from 3B-scale LLMs.

\section{Related Work}
Recent work has explored the use of large language models (LLMs) for post-ASR processing, including hypothesis rescoring, generative error correction, contextual correction, and speech-aware or training-time extensions.

\subsection{LLM-based ASR Hypothesis Rescoring and Generative Correction}

A growing body of work applies LLMs at inference time for post-ASR correction, framing the task as text generation from either a 1-best hypothesis or an N-best list. Prompting-based pipelines show that frozen LLMs can perform N-best rescoring or generative correction through in-context learning, including task-activating prompts for ASR correction~\cite{yangGenerativeSpeechRecognition2023}. The GenSEC challenge further formalizes this setting as a benchmark, covering transcription correction and related speech-processing tasks such as speaker tagging and emotion recognition~\cite{Yang_2024}.

Several works investigate how to better exploit N-best hypotheses. Some studies use LLMs to rewrite or correct from N-best lists, showing that generative correction can go beyond simple hypothesis reranking~\cite{ma2025asrerrorcorrectionusing}. Other works expand or refine the candidate space, for example by prompting LLMs to generate additional hypotheses before reranking them with acoustic and linguistic scores~\cite{tur2024progrespromptedgenerativerescoring}, or by using LLMs to select a smaller and more informative hypothesis subset before downstream correction~\cite{wuLLMDrivenHypothesisSet}. Prompt design has also been studied explicitly, with evolutionary prompt optimization used to search for more effective post-ASR correction instructions~\cite{sachdevEvolutionaryPromptDesign2024}.

Despite their effectiveness, these approaches typically rely on prompting frozen LLMs, often very large ones (e.g., GPT, LLaMA, BLOOM), and remain prone to over-editing, hallucinated corrections, and high computational cost. They also leave open how much of the desired select-or-rewrite behavior can instead be learned through fine-tuning, and at what model size such a fine-tuned post-editor becomes reliable. We study this question by comparing prompting against oracle-guided fine-tuning across model sizes.

\subsection{Robust, Contextual, and Multilingual ASR Error Correction}

Another line of work studies LLM-based ASR correction under more challenging or specialized conditions, including noisy speech, domain shift, rare words, named entities, and low-resource languages. Robustness-oriented methods incorporate additional signals such as N-best diversity, synthetic data, retrieval, and phonetic context to improve correction under noisy or domain-specific conditions~\cite{huLargeLanguageModels2024,ghosh2024failingforwardimprovinggenerative,yamashita2025llmbasedgenerativeerrorcorrection}. These studies show that correction quality depends not only on the LLM itself, but also on whether the input provides sufficient evidence for difficult words, named entities, and domain-specific terms.

Contextual ASR correction has also been explored in goal-oriented dialogue, where dialogue states, task context, and LLM-augmented task lists are used to rerank N-best hypotheses or recover plausible user goals~\cite{asano2025contextualasrerrorhandling}. This is related to our motivation in that it uses information beyond the Top-1 ASR output, but it focuses on task-specific contextual ranking rather than generative N-best post-editing. Unlike these methods, which improve correction mainly by enriching the input with extra signals or context, we keep the N-best text fixed and study which training examples are most useful for fine-tuning a generative post-editor.

Recent work further evaluates LLM-based correction across languages, tasks, and resource conditions. Multilingual and multitask benchmarks study generative correction for both ASR and speech translation across many languages and decoding settings~\cite{yang-etal-2025-covoger}, while low-resource studies show that model scale and adaptation strategy can strongly affect correction performance, with smaller fine-tuned models sometimes outperforming larger LLMs~\cite{kumar-etal-2026-post}.

\subsection{Speech-Aware and Training-Time Extensions}

While most LLM-based post-ASR correction methods operate on text hypotheses alone, recent work has explored adding speech or multimodal evidence to improve faithfulness to the source signal. Speech-aware and audio-visual approaches provide the LLM with source speech, speech-derived representations, or visual features in addition to N-best hypotheses, reducing the risk of fluent but acoustically unsupported corrections~\cite{huListenAgainChoose2024,liu2025listeningseeingagaingenerative}. These methods address a limitation of text-only post-editing, but require additional access to speech or multimodal inputs, whereas our setting focuses on N-best text hypotheses produced by an ASR system.

Other extensions use LLMs or specialized correction models beyond direct inference-time post-editing. Prior work has used textual and speech LLMs to refine hypotheses from multiple ASR systems for pseudo-label generation~\cite{prakashBetterPseudolabelingMultiASR2025}, explored compact task-specific correction models for better latency and controllability~\cite{guRevisitingASRError2026}, and studied full-text or 1-best correction with document-level, multilingual, or phonetic information~\cite{tangChainCorrectionFulltext2026,guRevisitingASRError2026}. These directions are complementary to our work. In contrast, we focus on reference-guided data selection for fine-tuning a generative N-best post-editor, asking whether a behavior-targeted subset can teach the model when to select, preserve, or cautiously rewrite hypotheses under a matched subset-size comparison.

\section{Method}
\label{sec:method}
\subsection{System Overview}
\begin{figure*}[t]
    \centering
    \includegraphics[width=\textwidth]{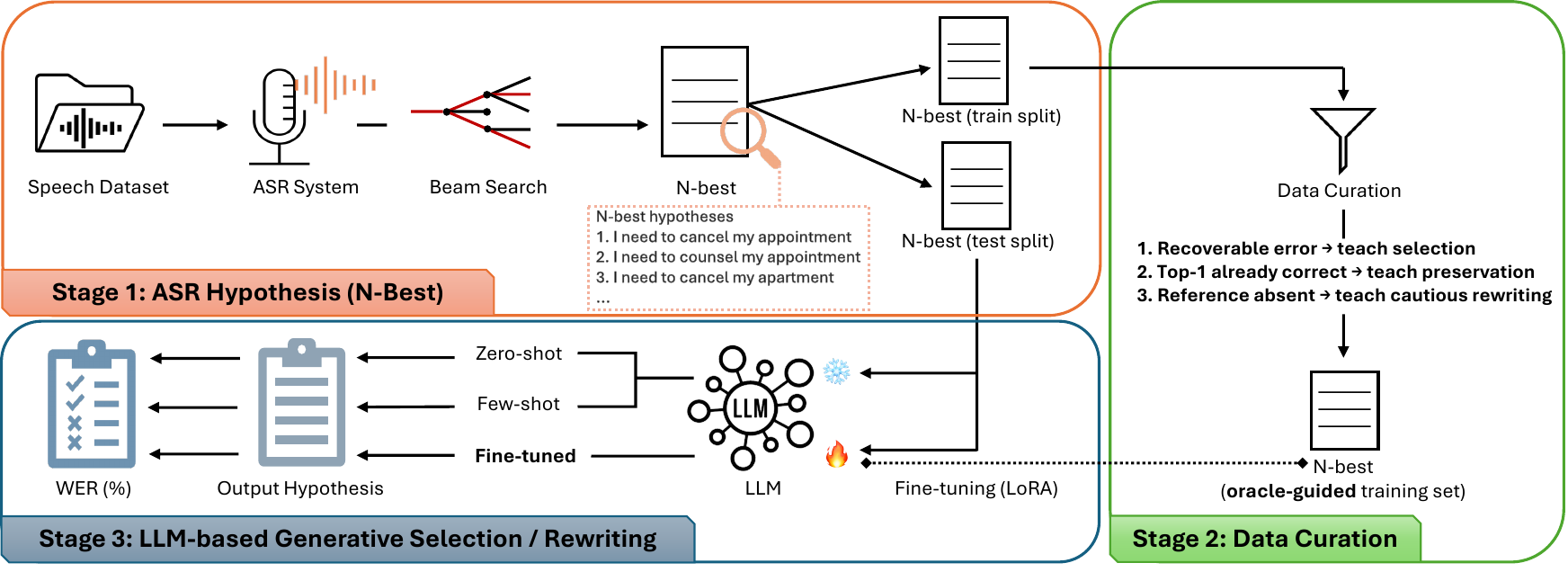}
    \caption{Pipeline overview of the oracle-guided select-or-rewrite framework. ASR beam search produces N-best lists for training and test utterances. Training N-best lists are compared with references and curated into behavior-targeted pools for LoRA fine-tuning, while test N-best lists are given to the LLM for generative selection or rewriting and evaluated with WER.}
    \label{fig:pipeline}
\end{figure*}

Our framework performs LLM-based generative rescoring of ASR outputs using N-best hypotheses, illustrated in Fig. \ref{fig:pipeline} with three stages, i.e., (1) ASR decoding with beam search, (2) Data curation, and (3) LLM-based select-or-rewrite post-editing. Formally, given an input speech signal $\mathbf{x}$, the ASR system generates a set of $N$ candidate hypotheses $\mathcal{H} = \{y_1, y_2, ..., y_N\}$. The goal is to identify the hypothesis $y^* \in \mathcal{H}$ or generate a new sequence $\hat{y}$ that minimizes the Word Error Rate (WER) with respect to the ground truth.

\noindent\textbf{Stage 1: ASR hypothesis generation.}
Specifically for this work, each input utterance is decoded using Whisper \cite{radfordRobustSpeechRecognition2023}, applying the beam search to produce an N-best list ($N=12$). The Top-1 hypothesis is the baseline ASR output, and the remaining candidates capture alternative transcriptions. This broader candidate set shifts disambiguation from the acoustic model toward language-level reasoning \cite{maNbestT5Robust2023}.

\noindent\textbf{Stage 2: Oracle-guided Data Curation and Fine-tuning.}
\label{sec:stage2}
Generic LLMs often over-correct or hallucinate on ASR outputs. We therefore build a balanced instruction-tuning dataset with oracle-guided curation.
We define three sample-level signals:
\begin{enumerate}
    \item \textbf{Difficulty ($\delta$):} WER of the Top-1 hypothesis.
    \item \textbf{Diversity Proxy ($\text{div}$):} mean pairwise WER between Top-1 and top-$K$ candidates.
    \item \textbf{Risk Flags:} indicators for numbers ($\mathbb{I}_{\text{num}}$), apostrophes ($\mathbb{I}_{\text{apos}}$), and proper nouns ($\mathbb{I}_{\text{prop}}$).
\end{enumerate}
The goal of this curation is not to maximize training-set coverage, but to construct a compact and behavior-targeted fine-tuning set. Positive corrections teach selection, regularizers teach preservation of correct Top-1 outputs, and hard negatives teach caution when the correct answer is not supported by the N-best list. Using these signals, we construct three training pools that enforce a hybrid policy (select when possible, rewrite when needed):
\begin{enumerate}
\item \textbf{Positive Corrections (Recoverable Errors):}
Top-1 is wrong, but the reference exists in the N-best list ($y_{\text{ref}} \in \mathcal{H}$). We retain 100\% of these samples to learn discriminative selection.
\item \textbf{Regularization Set (Anti-Forgetting):}
Top-1 is already correct ($y_{\text{top1}} = y_{\text{ref}}$). To discourage unnecessary edits, we keep the top 20\% by risk-weighted score:
\begin{equation}
S_{\text{reg}} =
2.0 \cdot \mathbb{I}_{\text{num}} +
1.0 \cdot \mathbb{I}_{\text{apos}} +
0.5 \cdot \mathbb{I}_{\text{prop}} +
0.5 \cdot \text{div} +
0.01 \cdot |y_{\text{ref}}|
\end{equation}
This emphasizes hard-but-correct examples (e.g., numbers and proper nouns).
\item \textbf{Hard Negatives (Hallucination Suppression):}
Reference is absent from the N-best list ($y_{\text{ref}} \notin \mathcal{H}$). We select subtle failures (not unintelligible noise) with $0.05 \le \delta \le 0.35$, using a ratio of 30\% relative to positive corrections.
\end{enumerate}

\noindent\textbf{Stage 3: Generative Listwise Rescoring.}
\label{sec:stage3}
We train the LLM on the oracle-curated data above so it learns two behaviors: (1) select a candidate from the list when recoverable evidence exists, and (2) rewrite when no candidate is sufficiently correct.

At inference, the LLM performs listwise select-or-rewrite post-editing: it receives the full N-best list at once and generates a single final transcript. Unlike pointwise rescoring, which scores hypotheses separately, this formulation allows the model to compare candidate differences as a proxy for acoustic ambiguity (e.g., phonetic neighbors), despite not directly observing audio $\mathbf{x}$. For each utterance, we provide all $K$ candidates ($K=N$) in a structured prompt and enforce format-compliant output to extract a single final hypothesis:

\begin{promptbox}
\ttfamily\footnotesize
You are an ASR error corrector. Below are $K$ candidate transcriptions for the same audio.\\
Candidates:\\
1. [Candidate 1] ... K. [Candidate K]\\
Task: Identify the correct sentence based on phonetic similarity and context. Output ONLY the corrected sentence.
\end{promptbox}

The LLM operates on the raw ASR hypotheses before text normalization. We keep the raw N-best candidates as input so that punctuation, casing, disfluencies, and other beam-search variations remain available to the model. The model is therefore fine-tuned and prompted using the original hypothesis text, and produces a raw post-edited transcript.

\section{Experimental Setup}
\subsection{Datasets}

We evaluate on two English datasets with complementary properties: \textit{Mozilla CommonVoice}, a public read-speech corpus, and \textit{definedAI}, a proprietary customer--agent conversational corpus. This choice allows us to compare LLM-based ASR post-editing across different domains, ASR difficulty levels, and degrees of public text exposure. In particular, \textit{CommonVoice} provides a widely used public benchmark with diverse speakers and possible overlap with publicly available text sources, whereas \textit{definedAI} provides a less-exposed, domain-specific evaluation setting with more controlled conversational content.

\begin{table}[ht]
\centering
\caption{Statistics of the two datasets used in this work.}
\label{tab:dataset_stats}
\begin{tabular}{lcccc}
\toprule
& \multicolumn{2}{c}{definedAI} & \multicolumn{2}{c}{CommonVoice} \\
\cmidrule(lr){2-3} \cmidrule(lr){4-5}
& Train & Test & Train & Test \\
\midrule
\# Utterances & 17398 & 2009 & 17000 & 16354 \\
Total Duration (hrs) & 38.17 & 4.35 & 27.98 & 27.05 \\
\bottomrule
\end{tabular}
\end{table}

\textbf{\textit{definedAI}\footnote{https://defined.ai}.}
A proprietary conversational corpus with relatively consistent and high-quality recording conditions, covering domains such as healthcare, telecommunications, retail, and banking. The data consists of customer--agent interactions, and we use its full train/test split.

\textbf{\textit{CommonVoice}.}
A large crowdsourced read-speech corpus with diverse speakers and recording environments~\cite{ardilaCommonVoiceMassivelyMultilingual2020}. The full English training split contains 948,736 utterances, corresponding to about 1505 hours of speech. For comparable computational cost and tractable N-best generation, we construct a speaker-balanced training subset by sampling 1,700 speakers and 10 utterances per speaker, resulting in 17,000 training utterances. The full \textit{CommonVoice} test set is retained.

For both datasets, fine-tuning examples are selected from the training split using the oracle-guided curation strategy described in Stage 2 of Section~\ref{sec:method}. Table~\ref{tab:dataset_stats_curation} summarizes the resulting curation funnel. Because \textit{CommonVoice} is public read speech while \textit{definedAI} is proprietary conversational speech, we return to the implications of this dataset difference in Section~\ref{sec:discussion}.

\begin{table*}[ht!]
\centering
\caption{\textbf{Statistics of the Oracle-Guided Fine-tuning Dataset}. Recoverable errors ($y_{\text{ref}}\in\mathcal{H}$ and $y_{\text{top1}}\neq y_{\text{ref}}$) are fully retained. Correct predictions (regularizers) are downsampled by keeping the top 20\% according to a risk-weighted ranking score. Additional hard negatives ($y_{\text{ref}}\notin\mathcal{H}$) are sampled at a ratio of 0.3 relative to the number of recoverable errors, filtered by requiring the top-1 WER to lie in $[0.05,0.35]$ and the N-best diversity score is set to 0.08. Finally, 8\% of the curated samples are reserved as validation using a stratified split over sample category and difficulty bins.}
\label{tab:dataset_stats_curation}

\begin{tabular}{lcccccc c cccccc}
\toprule

& \multicolumn{6}{c}{Before Filtering} 
& 
& \multicolumn{6}{c}{After Filtering} \\

\cmidrule(lr){2-7} \cmidrule(lr){9-14}

Category (Training Set)
& \shortstack{DAI\\W-S} 
& \shortstack{DAI\\W-M} 
& \shortstack{DAI\\W-L}
& \shortstack{CV\\W-S} 
& \shortstack{CV\\W-M}
& \shortstack{CV\\W-L}
& $\rightarrow$
& \shortstack{DAI\\W-S} 
& \shortstack{DAI\\W-M} 
& \shortstack{DAI\\W-L}
& \shortstack{CV\\W-S} 
& \shortstack{CV\\W-M} 
& \shortstack{CV\\W-L} \\

\midrule

Positive (Recoverable)
& 2485 & 2512 & 2413 & 2897 & 2772 & 2694
& $\rightarrow$
& 2286 & 2311 & 2221 & 2665 & 2550 & 2479 \\

Regularizers (Correct)
& 5994 & 6387 & 6419 & 9319 & 10447 & 10723
& $\rightarrow$
& 1102 & 1175 & 1180 & 1714 & 1923 & 1972 \\

Hard Negatives (Hopeless)
& 8919 & 8499 & 8566 & 4784 & 3781 & 3583
& $\rightarrow$
& 686 & 693 & 665 & 800 & 764 & 743 \\

\midrule

Validation (8\%)
&  &    &  &    &   & 
& $\rightarrow$
& 354 & 363 & 353 & 450 & 455 & 452 \\

\midrule

\textbf{Total}
& 17398 & 17398 & 17398 & 17000 & 17000 & 17000
& $\rightarrow$
& 4074 & 4179 & 4066 & 5179 & 5237 & 5194 \\

\bottomrule
\end{tabular}

\end{table*}

\subsection{Models}

We use \textbf{Whisper-Small} (244M), \textbf{Whisper-Medium} (769M), and \textbf{Whisper-Large} (1550M) as ASR frontends~\cite{radfordRobustSpeechRecognition2023}. For each utterance, we generate N-best hypotheses with beam size 12 using HuggingFace Transformers. For post-editing, we use \textbf{Qwen2.5-Instruct} models~\cite{qwenQwen25TechnicalReport2025} at \textbf{0.5B, 3B, 7B, and 14B} parameters to study scale effects.

For ASR inference, LLM inference, and LLM fine-tuning, we use an NVIDIA RTX 3090 GPU with 24GB memory. The only exception is Qwen-14B, which requires an NVIDIA H100 GPU with 80GB memory for both inference and fine-tuning.

For fine-tuning, we use \textbf{LoRA}~\cite{huLoRALowRankAdaptation2021a} with the \textbf{Unsloth} framework, keeping the base model weights frozen. Training uses full precision without quantization, 2 epochs, and an effective batch size of 8. LoRA uses rank $r=64$, $\alpha=16$, and dropout 0.05 on all linear layers.

\subsection{Baselines and Comparisons}

The primary ASR baseline is the Top-1 hypothesis from the Whisper N-best list, corresponding to the highest-scoring output produced by beam-search decoding. This represents the standard ASR output without any LLM-based post-editing.

We further compare against two prompting-based LLM baselines: zero-shot prompting and few-shot prompting. In both settings, the frozen LLM receives the full N-best candidate list and generates a corrected transcription using the prompt described in Section~\ref{sec:stage3}. These baselines evaluate whether a frozen LLM can perform generative N-best post-editing without task-specific fine-tuning.

For \textit{few-shot prompting}, we use four in-context examples. To ensure comparability across ASR frontends and model scales, the demonstration set is fixed for each dataset. Specifically, for all Whisper-Small, Whisper-Medium, and Whisper-Large outputs on \textit{definedAI}, the demonstrations are sampled once from the \textit{definedAI + Whisper-Small} training data and reused across experiments. The same protocol is applied to \textit{CommonVoice}. This avoids introducing additional variability from changing few-shot examples across ASR systems. Few-shot prompting uses the same instruction as zero-shot prompting, with four demonstrations prepended before the test case:

\begin{promptbox}
\ttfamily\footnotesize
\#\#\# Example i\\
Candidates:
1. ...
...
N. ...\\
Correct Output: ...

\#\#\# Test case\\
Candidates:
1. ...
...
N. ...\\
Correct Output:
\end{promptbox}

To evaluate the proposed \textit{oracle-guided fine-tuning}, we fine-tune the LLM on the curated training subset constructed by the data selection strategy in Section~\ref{sec:stage2}. The resulting model performs generative selection or rewriting from the full N-best list.

To isolate the effect of data curation, we introduce a \textit{random-data fine-tuning baseline}. This baseline uses the same fine-tuning procedure and the same number of training examples as the oracle-guided setting, but samples are drawn randomly from the original training set without oracle-based filtering. The validation and test sets are kept identical across fine-tuning experiments. Therefore, the comparison measures whether the proposed curation strategy provides more effective supervision than a random subset under a matched fine-tuning data budget.

\subsection{Evaluation}
We evaluate system performance using Word Error Rate (WER, \%), the standard metric for automatic speech recognition. While the LLM operates on raw ASR hypotheses, WER is computed after normalization is applied to the text.

\section{Results}
\begin{figure}[t]
    \centering
    \includegraphics[width=1\linewidth]{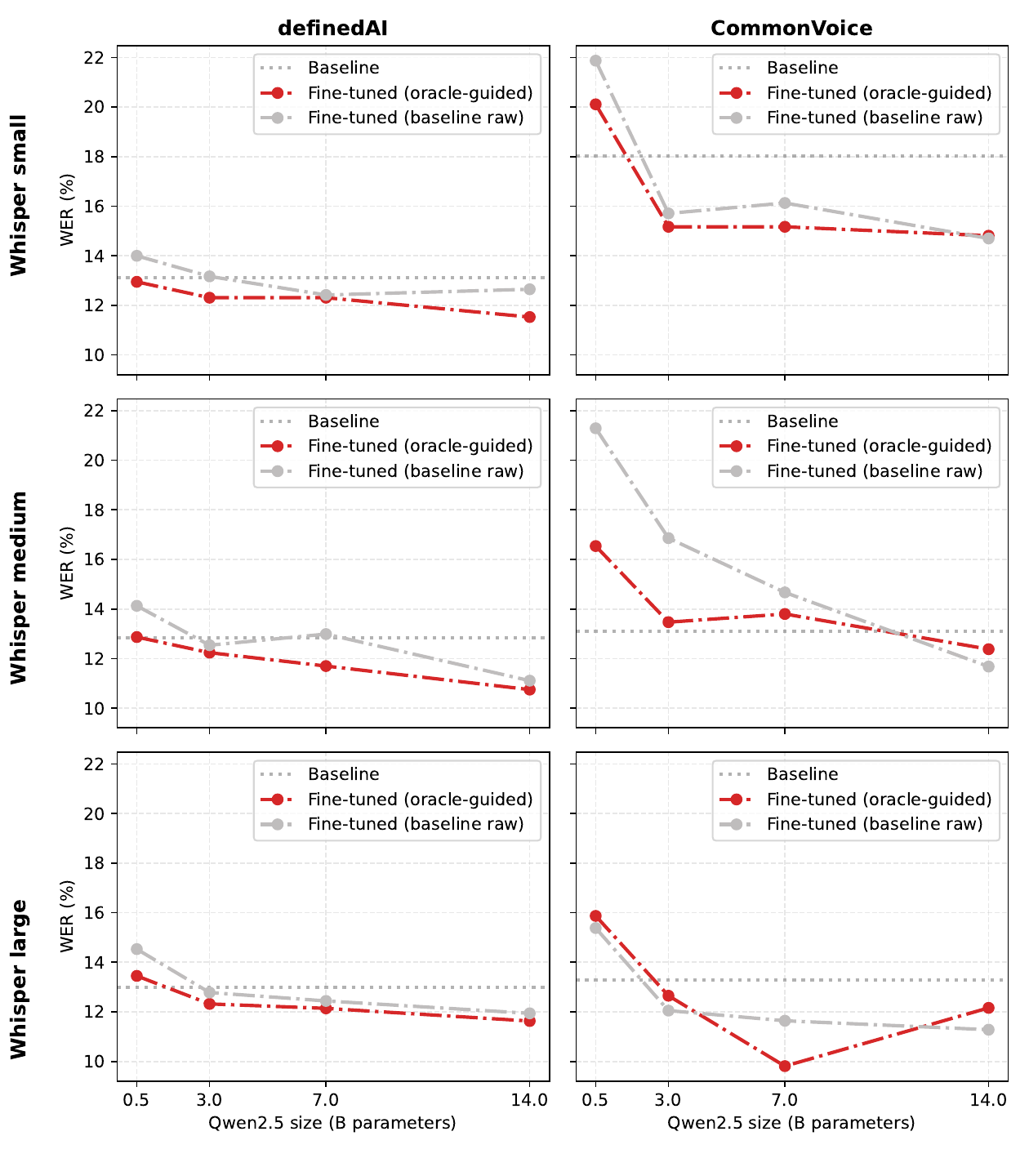}
    \caption{Comparison between oracle-guided and random-data fine-tuning under a matched training-data budget. The random baseline follows the same fine-tuning setup and uses the same number of examples as the oracle-guided setting, but samples them randomly from the training split without oracle-based curation. Results are shown for Whisper-Small, Whisper-Medium, and Whisper-Large on \textit{definedAI} and \textit{CommonVoice}.}
    \label{fig:performance_oracle-baseline}
\end{figure}

\begin{figure}[t]
    \centering
    \includegraphics[width=1\linewidth]{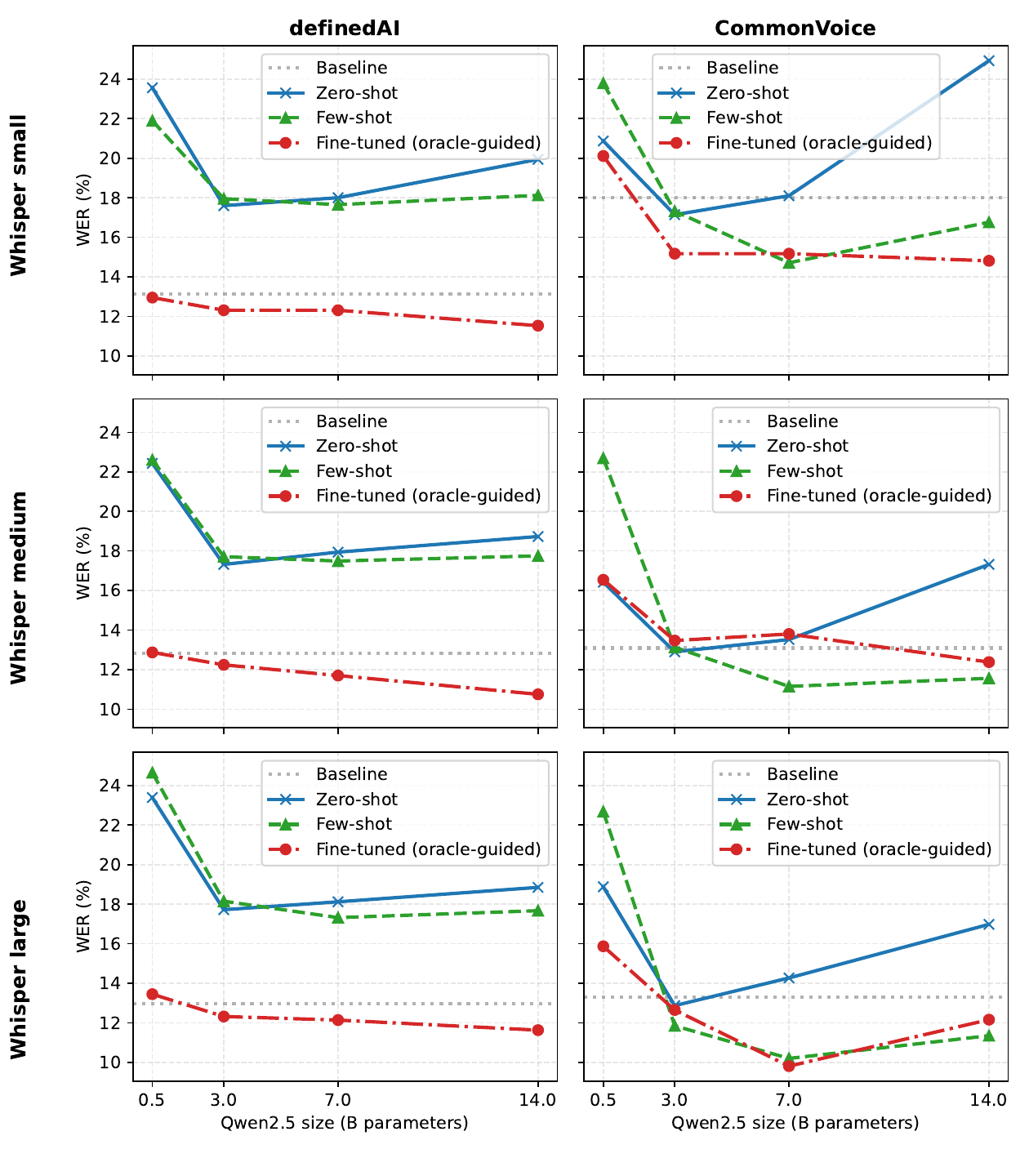}
    \caption{Comparison of zero-shot prompting, few-shot prompting, and oracle-guided fine-tuning for LLM-based ASR post-editing. Results are shown across Qwen2.5 model sizes using Whisper-Small, Whisper-Medium, and Whisper-Large ASR frontends on \textit{definedAI} and \textit{CommonVoice}.}
    \label{fig:performance_zero-few-oracle}
\end{figure}

\subsection{Effect of Oracle-Guided Data Curation}

Figure~\ref{fig:performance_oracle-baseline} compares oracle-guided fine-tuning with random-data fine-tuning under a matched training-data budget. On \textit{definedAI}, the gains are modest but consistent in direction: from around 3B parameters, oracle-guided fine-tuning brings WER to or below the ASR baseline and is slightly better than or on par with random-data fine-tuning across the three Whisper backbones, whereas at 0.5B fine-tuning is unstable and can perform worse than the baseline. This suggests that, under a matched budget, the curated subset is at least as effective as randomly selected examples, with a small but generally consistent advantage from 3B upward.

On \textit{CommonVoice}, oracle-guided fine-tuning improves performance in most settings, but its advantage over random-data fine-tuning is less consistent. This suggests that the effectiveness of data curation depends on dataset properties. We also observe that 0.5B models are often insufficient for reliable fine-tuning, while improvements become clearer from around 3B parameters, especially on the proprietary \textit{definedAI} dataset.

\begin{figure}[ht]
\centering
\begin{tcolorbox}[
    enhanced,
    width=\columnwidth,
    colback=white,
    colframe=black!25,
    colbacktitle=black!12,
    coltitle=black,
    title={Example: LLM selecting the correct raw hypothesis from the n-best list},
    fonttitle=\bfseries\footnotesize,
    boxrule=0.45pt,
    arc=2pt,
    left=5pt,
    right=5pt,
    top=4pt,
    bottom=4pt,
    toptitle=3pt,
    bottomtitle=3pt,
    titlerule=0pt
]
\fontsize{7.2}{8.4}\selectfont
\setlength{\tabcolsep}{0pt}
\renewcommand{\arraystretch}{1.18}

\begin{tabularx}{\linewidth}{
@{}>{\bfseries\raggedright\arraybackslash}p{0.20\linewidth}
   >{\raggedright\arraybackslash}X@{}
}
Reference &
perfect thanks a lot for that again thank you \\

Raw n-best hypotheses &
\begin{minipage}[t]{\linewidth}
\vspace{-1.5ex}
\begin{enumerate}[leftmargin=1.15em,itemsep=0.5pt,topsep=0pt]
    \item Perfect. \textcolor{red}{\textbf{Things are up with}} that again. Thank you.
    \item Perfect, \textcolor{red}{\textbf{things are up with}} that again. Thank you.
    \item Perfect. \textcolor{blue}{\textbf{Thanks a lot for}} that again. Thank you.
    \item Perfect, \textcolor{blue}{\textbf{thanks a lot for}} that again. Thank you.
    \item Perfect. \textcolor{red}{\textbf{Things were up with}} that again. Thank you.
    \item Perfect. \textcolor{red}{\textbf{Things are up without a game.}} Thank you.
    \item Perfect, \textcolor{red}{\textbf{things are up with}} that again, thank you.
    \item Perfect. \textcolor{red}{\textbf{Things are up without}} again. Thank you.
\end{enumerate}
\vspace{-1.5ex}
\end{minipage}
\\
\\
LLM output &
perfect \textcolor{blue}{\textbf{thanks a lot for}} that again thank you\\
\end{tabularx}
\end{tcolorbox}

\vspace{-1.5mm}
\caption{Example of LLM-based hypothesis selection using raw ASR hypotheses. The Top-1 hypothesis contains a lexical substitution, where ``thanks a lot for'' is misrecognized as ``things are up with''. The correct phrase appears in lower-ranked N-best candidates, and the LLM selects it to produce the normalized output. This illustrates recoverable-error correction through listwise comparison of raw beam-search hypotheses.}
\label{fig:qualitative_selection_raw_example}
\end{figure}

\subsection{Prompting vs. Fine-tuning}

Figure~\ref{fig:performance_zero-few-oracle} compares zero-shot prompting, few-shot prompting, and oracle-guided fine-tuning. On \textit{definedAI}, prompting performs poorly across all Whisper backbones and LLM sizes, with both zero-shot and few-shot results remaining above the ASR baseline. In contrast, oracle-guided fine-tuning consistently reduces WER below the ASR baseline, indicating that task-specific adaptation is important for generative post-editing on this dataset.

On \textit{CommonVoice}, the trend is different. Prompting already improves over the ASR baseline in several settings, and few-shot prompting can match or outperform oracle-guided fine-tuning for some model sizes and ASR backbones. This shows that the relative benefit of fine-tuning over prompting is dataset-dependent. We discuss possible reasons for this interesting dataset-dependent behavior in Section~\ref{sec:discussion}.




\section{Discussion}
\label{sec:discussion}
\subsection{Effect of Behavior-Targeted Curation}

The WER reduction after fine-tuning shows that oracle-guided curation provides useful supervision for generative ASR post-editing. Under a matched data budget, the curated subset is more effective than randomly sampled fine-tuning data, suggesting that the selected examples are informative for teaching the LLM when to select from the N-best list and when to rewrite. This supports the main motivation of our method: instead of using arbitrary training examples, we construct a compact subset that targets the key post-editing behaviors, including recovering errors, preserving already-correct hypotheses, and avoiding unsupported rewrites.

The difference between \textit{definedAI} and \textit{CommonVoice} further suggests that the effect of such fine-tuning depends on the evaluation dataset. We discuss two possible factors: dataset domain and potential data exposure.

\subsection{Dataset Domain}

One possible reason for the dataset-dependent results is the difference in domain and text distribution. \textit{definedAI} is a proprietary corpus from specific application domains, such as healthcare, telecommunications, retail, and banking with turn-taking conversations between customer and agent. Fine-tuning on this data may therefore help the LLM adapt to a more specific target distribution and learn how to use N-best hypotheses for selection and rewriting in that setting. In contrast, \textit{CommonVoice} is a public read-speech corpus with broader and more diverse text. Prompting may already be competitive in this setting, reducing the relative benefit of task-specific fine-tuning.

\subsection{Potential Data Exposure}

Another possible factor is data exposure of the benchmark text. \textit{CommonVoice} includes speech prompts collected from public-domain text sources, including sources such as Wikipedia text \cite{ardilaCommonVoiceMassivelyMultilingual2020}. Since the pretraining corpora of modern LLMs are not fully transparent, we cannot rule out overlap between such benchmark text and the text seen during LLM pretraining. Therefore, performance on public read-speech benchmarks may not purely reflect the model's ability to use ASR N-best hypotheses for generative selection and rewriting; it may also be influenced by prior textual familiarity.

This less conclusive behavior on \textit{CommonVoice} raises a caution for evaluating LLM-based ASR post-editing on public test sets, especially when the benchmark text may have been exposed during pretraining. Evaluation on proprietary, domain-specific, or otherwise less-exposed datasets is therefore important for assessing whether the model has learned genuine post-editing behavior.

\subsection{Model-Scale Effect}

Besides dataset effects, model size also influences performance. The 0.5B model is often unstable and does not consistently improve over the ASR baseline, suggesting that very small LLMs may lack sufficient capacity for reliable listwise comparison and rewriting. In contrast, models around 3B parameters already show clear gains after oracle-guided fine-tuning. Larger models can further reduce WER in some settings, but the improvement is relatively small compared with the increase in model size. This makes the 3B scale a practical operating point for data-efficient generative ASR post-editing.

\section{Conclusion}
\label{sec:conclusion}

We presented an oracle-guided fine-tuning framework for generative ASR post-editing from N-best hypotheses. The proposed curation strategy constructs behavior-targeted supervision from recoverable errors, already-correct regularizers, and hard negatives, teaching the LLM when to select from the N-best list, when to preserve the Top-1 hypothesis, and when to avoid unsupported rewriting. Under a matched fine-tuning data budget, the curated subset provides effective supervision, with the clearest gains on the proprietary \textit{definedAI} dataset and reliable improvements emerging from around 3B LLM parameters. Results on \textit{CommonVoice} are less conclusive: prompting is already competitive, and possible benchmark-text exposure from public read-speech material may affect the interpretation of results. These findings suggest that reference-guided curation can support compact and behavior-targeted ASR post-editing, while also highlighting the need to evaluate LLM-based post-editing on proprietary, domain-specific, or otherwise less-exposed datasets.



\vspace{12pt}

\bibliographystyle{IEEEtran}
\bibliography{refs}

\end{document}